\documentclass{article}

\usepackage[preprint]{corl_2025} 

\usepackage{graphicx}
\usepackage{amsmath}
\usepackage{amssymb}
\usepackage{array}
\usepackage[capitalise]{cleveref}
\Crefname{lstlisting}{Listing}{Listings}
\usepackage{listings}

\title{Residual Neural Terminal Constraint for MPC-based Collision Avoidance in Dynamic Environments}

\author{
  Bojan Derajić$^{1}$,
  Mohamed-Khalil Bouzidi$^{2, 3}$,
  Sebastian Bernhard$^{2}$ and
  Wolfgang Hönig$^{1}$ \\
  $^{1}$Technical University of Berlin, Germany \\
  $^{2}$Continental Automotive Technologies GmbH, Germany \\
  $^{3}$Free University of Berlin, Germany
}

\begin{document}
\maketitle


\begin{abstract}
    In this paper, we propose a hybrid MPC local planner with a learning-based approximation of a time-varying safe set, which is derived from local observations and applied as the terminal constraint. This set can be represented as the zero-superlevel set of the value function computed via Hamilton-Jacobi (HJ) reachability analysis, which is often infeasible in real-time. We exploit the property that the HJ value function can be expressed as a difference of the corresponding signed distance function (SDF) and a non-negative residual function. The residual component is modeled as a neural network with non-negative output and subtracted from the computed SDF, resulting in a real-time value function estimate that is at least as safe as the SDF by design. Additionally, we parametrize the neural residual by a hypernetwork to improve real-time performance and generalization properties. The proposed method is compared with three state-of-the-art methods in simulations and hardware experiments, achieving up to 30\% higher success rates compared to the best baseline while requiring a similar computational effort and producing high-quality (low travel-time) solutions.
\end{abstract}



\section{Introduction}
\label{sec:intro}
	
As robots increasingly operate in complex and unstructured settings alongside humans and other moving agents, the ability to achieve safe collision avoidance while maintaining operational efficiency becomes crucial. This challenge is particularly emphasized when robots have limited actuation and must navigate through unknown environments based only on sensor observations. In many examples, relying on other agents to avoid collision might be completely unsafe. For example, a cleaning robot in a public space should prevent collisions with humans driving a segway, e-scooter or a skateboard since they move relatively fast and have limited steering capabilities. Similarly, an intralogistics robot in a warehouse should give priority to human-operated forklifts transporting heavy and unstable objects. Although those scenarios might appear simplistic, they become challenging for nonholonomic mobile robots with limited actuation \cite{hwang_safe_2024, fraichard_short_2007}. 

One of the most prominent techniques for motion planning and control in mobile robotics is model predictive control (MPC), which consists of a few core components such as a robot model, cost function, constraints, terminal set, and optimization algorithm \cite{mayne_constrained_2000}. Even though a proper choice of any component is not trivial, designing a terminal set that is control invariant is particularly difficult, especially in dynamic environments. This set is necessary to achieve the recursive feasibility of the MPC, which is a crucial property in safety-critical applications.

To determine the maximal control invariant set, which we call \textit{maximal safe set}, Hamilton-Jacobi (HJ) reachability analysis can be used. This method relies on the dynamic programming principle to solve for the HJ value function and often cannot be executed online. To overcome this issue, we propose a learning-based approach to approximate the value function based on local observations. Moreover, we exploit the property that the HJ value function is always less than or equal to the corresponding signed distance function (SDF). This allows us to estimate only the residual function using a neural model with non-negative output and subtract it from the numerically computed SDF. In this way, the learned HJ value function approximation is guaranteed, by design, to be at least as safe as the corresponding SDF. The main contributions of this paper are:
\begin{itemize}
    \item A learning-based method for real-time estimation of the maximal safe set in dynamic environments with a model architecture that guarantees, by design, the same or higher level of safety compared to an SDF-based approach.
    \item Extensive empirical comparison of the proposed method with state-of-the-art methods in simulations and on a physical robot.
\end{itemize}


\section{Related Work}
\label{sec:related_work}

Robot navigation often requires some assumptions about the environment in which the robot is deployed. If the environment is shared with other agents, a standard distinction is whether the agents are interactive or not. In the first case, the robot and other agents collaboratively avoid collisions, while in the second case, the robot treats other agents as non-interactive moving obstacles. Some classical algorithms for interactive planning, such as ORCA \cite{van_den_berg_reciprocal_2011}, are based on the velocity obstacle (VO) paradigm \cite{fiorini_motion_1998}. The ORCA method is also combined with an MPC planner to formulate velocity constraints \cite{cheng_decentralized_2017}, predict human motion \cite{samavi_sicnav_2025}, or refine diffusion-based motion prediction \cite{samavi_sicnav-diffusion_2025}. Another popular approach is to learn interaction-aware motion prediction and then integrate it within the MPC planner as an environment model \cite{zhu_learning_2021}.

As indicated in \cref{sec:intro}, our work is focused on non-interactive scenarios where the assumption about reciprocal collision avoidance might lead to an unsafe situation. Similar to the interactive case, methods used in such environments are usually based on distance functions \cite{oleynikova_signed_2016} or velocity obstacles \cite{fiorini_motion_1998}. Focusing on MPC-based planners, one can use SDF \cite{zhang_optimization-based_2021} or VO \cite{piccinelli_mpc_2023} to define collision avoidance constraints. Another approach is introduced in \cite{zeng_safety-critical_2021}, where the authors combine MPC with a discrete-time control barrier function (CBF). However, none of these methods explicitly consider terminal set design and rely on a sufficiently long prediction horizon for improved recursive feasibility. On the other hand, our aim is to design a control invariant set in real-time and employ it as the terminal set for the MPC planner.

The maximal control invariant set can be determined using HJ reachability analysis, which is a framework for model-based computation of reachable sets~\cite{bansal_hamilton-jacobi_2017}. This framework suffers from the curse of dimensionality and is impractical for time-critical systems, which motivated methods such as efficient initialization, system decomposition, and neural approximation of the HJ value function \cite{herbert_reachability-based_2019, chen_decomposition_2018, bansal_deepreach_2021}. However, those approaches assume complete \textit{a priori} knowledge about the environment, which is often not satisfied, especially for dynamic environments. The work in \cite{borquez_parameter-conditioned_2023} introduced parameter-conditioning by augmenting the system dynamics with a set of ``virtual'' states, and the idea was applied in \cite{jeong_parameterized_2024} and \cite{nakamura_online_2023} to motion planning tasks. The main limitation of this method is that the dimensionality of the problem increases with the number of parameters. To overcome this, authors in \cite{derajic_learning_2025} propose a learning-based HJ value function estimation based on high-dimensional environment observations. Yet, this method was applied only to static environments without any formal guarantees on the approximation error. In our work, we extend this idea to dynamic environments and propose a new model architecture that guarantees that the value function estimate is at least as safe as the SDF.  


\section{Preliminaries}
\label{sec:preliminaries}

\subsection{Problem Formulation}
\label{subsec:problem_formulation}

We consider a nonlinear dynamical system
\begin{equation} \label{eq:general_dynamics}
    \dot{x}(t) = f(x(t), u(t)),
\end{equation}
where $x(t) \in \mathcal{X} \subseteq \mathbb{R}^n$ represents the system's state vector, $u(t) \in \mathcal{U} \subseteq \mathbb{R}^m$ is the control vector, and $t \in \mathbb{R}$ denotes the time variable. We are also interested in the discrete-time form of \cref{eq:general_dynamics}
\begin{equation}
    x_{k+1} = f_d(x_k, u_k),
\end{equation}
where $x_k$ and $u_k$ denote state and control vectors at discrete time $k \in \mathbb{Z}$, and $f_d$ is a discrete-time model obtained from $f$ with sampling time $\delta t$.

The MPC local planner solves a finite-horizon optimal control problem in a receding horizon manner, which is for time $t = k \cdot \delta t$ defined as follows:
\begin{subequations}
\begin{align}
    \min_{\substack{x_{k+1:k+N} \\ u_{k:k+N-1}}} \quad &\ell_N(x_{k + N}) + \sum_{i = 0}^{N - 1} \ell(x_{k+i}, u_{k+i}) \label{eq:opt_prob} \\
    s.t. \quad &x_{k+i+1} = f_d(x_{k+i}, u_{k+i}), \, i=0:N-1 \label{eq:dynamics_constr} \\
    &x_{k+i} \in \mathcal{X}, \quad i=1:N \label{eq:state_constr} \\
    &u_{k+i} \in \mathcal{U}, \quad i=0:N-1 \label{eq:control_constr} \\
    &h_{k+i}(x_{k+i}) \geq 0, \quad i=1:N-1 \label{eq:stage_constr} \\
    &h_{k+N}(x_{k+N}) \geq 0. \label{eq:term_constr}
\end{align}
\end{subequations}
In the formulation above, $x_{k+i}$ and $u_{k+i}$ are predicted state and control vectors at future time $k+i$ made at the current time step $k$ over the prediction horizon of length $N$. Constraint \cref{eq:dynamics_constr} enforces the system's dynamics, \cref{eq:state_constr} and \cref{eq:control_constr} ensure that the predicted state and control vectors have admissible values, while \cref{eq:stage_constr} and \cref{eq:term_constr} represent stage and terminal constraints, respectively. For the stage cost $\ell(x, u)$ and terminal cost $\ell_N(x)$ we use the standard quadratic form: 
\begin{equation} \label{eq:q}
    \ell(x, u) = (x - x_r)^\top Q (x - x_r) + u^\top R u, 
\end{equation}
\begin{equation} \label{eq:p}
    \ell_N(x) = (x - x_r)^\top Q_N (x - x_r),
\end{equation}
where $x_r$ is the reference state and $R$, $Q$ and $Q_N$ are weight matrices. In this work, we focus our attention on the terminal constraint function in \cref{eq:term_constr} and the main goal is to design $h_{k+N}(x_{k+N})$ in real-time such that its zero-superlevel set represents the maximal safe set.

\subsection{Hamilton-Jacobi Reachability Analysis in Time-Varying Environments}
\label{subsec:hj_reachability}

HJ reachability analysis is a model-based framework for formal verification of dynamical systems. It provides a method to compute a backward reachable tube (BRT) — the set of states such that if a system's trajectory starts from this set, it will eventually end up inside some failure set $\mathcal{F}(t)$ \cite{bansal_hamilton-jacobi_2017, fisac_reach-avoid_2015}. In the context of local motion planning, the BRT represents regions around obstacles where collisions become unavoidable (e.g., due to long braking distances, limited steering, or short prediction horizons).

Considering a system described by \cref{eq:general_dynamics} and assuming that the system starts at state $x(t)$, we denote with $\xi^{u}_{x, t}(\tau)$ the system's state at time $\tau$ after applying $u(\cdot)$ over time horizon $[t, \tau]$. The BRT from which the system will reach the failure set $\mathcal{F}$ within the time horizon $[t, T]$ is defined as
\begin{equation}
    \mathcal{B}(t) = \{ x: \forall u(\cdot), \, \exists \tau \in [t, T], \, \xi^{u}_{x, t}(\tau)  \in \mathcal{F}(\tau) \}.
\end{equation}

The set $\mathcal{F}(t)$ is defined as the zero-sublevel set of a failure function $F(x, t)$ (SDF in our context), i.e. ${\mathcal{F}(t) = \{x:F(x, t) \leq 0 \}}$, and to compute the BRT, we define cost function
\begin{equation}
    J(x, u(\cdot), t) = \min_{\tau \in [t, T]} F(\xi^{u}_{x, t}(\tau), \tau).
\end{equation}
Since $\mathcal{F}(t)$ represents occupied space at time $t$, the goal is to find the optimal control that will maximize the cost $J$. Therefore, we introduce the value function
\begin{equation} \label{eq:value_func}
    V(x, t) = \sup_{u(\cdot)} \{ J(x, u(\cdot), t) \}.
\end{equation}
The value function is computed over the system state space by solving the following Hamilton-Jacobi-Bellman Variational Inequality (HJB-VI) using the dynamic programming principle \cite{barron_bellman_1989, fisac_reach-avoid_2015}:
\begin{equation} \label{eq:hjb_vi}
    \begin{aligned}
    \min \left\{ \frac{\partial}{\partial t}V(x, t) + H(x, t), \, F(x, t) - V(x, t) \right\} &= 0, \\
    V(x, T) &= F(x, T).
    \end{aligned}
\end{equation}
In the formulation above, Hamiltonian $H(x, t)$ is defined as
\begin{equation}
    H(x, t) = \max_{u \in \mathcal{U}} \nabla V(x, t)^\top \cdot f(x, u).
\end{equation}
Once the value function \cref{eq:value_func} is computed, the corresponding BRT can be obtained as its zero-sublevel set, i.e.
\begin{equation}
    \mathcal{B}(t) = \left\{ x: V(x, t) \leq 0 \right\},
\end{equation}
and the maximal safe set $\mathcal{S}(t)$ is its complement: $\mathcal{S}(t) = \mathcal{B}(t)^c$.


\section{Methodology}
\label{sec:methodology}

In this section, we present our framework for online estimation of the maximal safe set in unknown dynamic environments. The goal is to estimate this set in real-time and integrate it within the MPC local planner as the terminal constraint. As explained in \cref{subsec:hj_reachability}, the maximal safe set at some time $t'$ is the zero-superlevel set of the HJ value function $V(x, t)$ and it depends on the evolution of the failure set $\mathcal{F}(\tau)$ for $\tau > t$. If we consider the MPC problem formulation defined for time $k \cdot \delta t$, we need to estimate the maximal safe set at time ${(k+N) \cdot \delta t}$, i.e. the value function $V(x, (k+N) \cdot \delta t)$. To do so, it is necessary to reason about the evolution of the environment and the motion of the other agents even after the prediction horizon. This setting is very different from static environments considered in \cite{derajic_learning_2025}, where the failure set does not change over time. In this case, we need to introduce a certain assumption about the motion of the agents even after the final time step. In this paper, it is assumed that the moving obstacles will continue to move with constant velocity, but other assumptions can be adopted as well (see \cref{sec:limitations}). This allows us to condition the safe set estimation at time ${(k+N)\cdot\delta t}$ based on predictions over the time horizon ${[k\cdot\delta t, (k+N)\cdot\delta t]}$.

To achieve real-time performance, we rely on the idea introduced in \cite{derajic_learning_2025} where the model is separated into two neural networks --- a hypernetwork and a main network. The hypernetwork is an expressive model with trainable parameters that parametrizes the main network based on the conditioning input. Complementary, the main network is a lightweight model that is parametrized by the hypernetwork and integrated within numerical optimization as a constraint function. In our approach, we use the tail of the SDF sequence, corresponding to the predicted motion of the surrounding agents, as the input to the hypernetwork. At time step $k$, we obtain predicted positions ${\{ p_{k}^j, p_{k+1}^j, \ldots, p_{k+N}^j \}_{j=1}^M}$ for $M$ agents over the prediction horizon and compute the corresponding SDF sequence ${\{ F_{k}(x), F_{k+1}(x), \ldots, F_{k+N}(x) \}}$. The first $N$ SDFs are used as the stage constraints, i.e. $h_{k+i}(x) = F_{k+i}(x)$ for $i=0,1,\ldots, N-1$, and the last $K$ SDFs, from $N-K+1$ to $N$, are used as the input to the hypernetwork to generate main network parameters $\Theta$.

The method introduced in \cite{derajic_learning_2025} uses the main network to directly approximate the value function $V_{k+N}(x) = V(x, (k+N) \cdot \delta t)$ at the end of the horizon. However, from the HJB-VI defined by \cref{eq:hjb_vi} we can conclude that $V_{k+N}(x)$ is always less than or equal to $F_{k+N}(x), \forall x$. In other words, $V_{k+N}(x)$ can be represented as a difference of $F_{k+N}(x)$ and a non-negative residual $R_{k+N}(x)$, i.e.
\begin{equation}
    V_{k+N}(x) = F_{k+N}(x) - R_{k+N}(x) \quad s.t. \quad R_{k+N}(x) \geq 0, \forall x.
\end{equation}
This observation allows us to approximate only the residual $R_{k+N}(x)$ by the main network and reuse the numerically computed SDF at time $k+N$. Moreover, by applying a non-negative activation function at the output of the main network, we are guaranteed, by design, that the estimated residual is always non-negative for any input value $x$. Consequently, all states that are classified as unsafe by $F_{k+N}(x)$ are also guaranteed to be classified as unsafe by the resulting value function estimate $\hat{V}_{k+N}(x)$, which is important from a safety perspective. In addition, as we illustrate through an ablation study in \cref{apx:ablation_study}, the learning task is simplified since the model needs to learn only the difference between the SDF and the value function, instead of learning the value function directly without any inductive bias.

\begin{figure}[tb]
    \centering
    \includegraphics[width=1\linewidth]{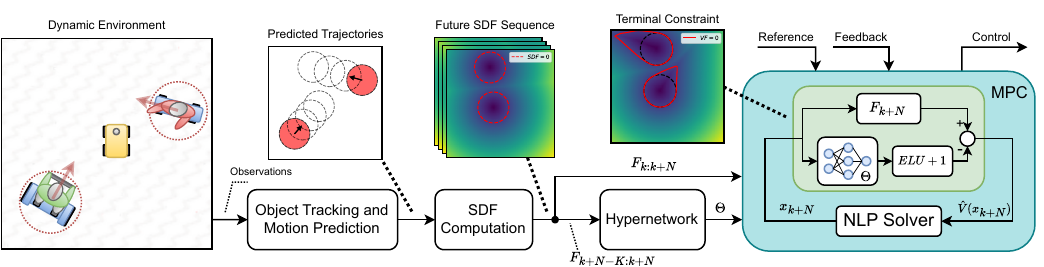}
    \caption{The proposed RNTC-MPC framework. Based on tracked positions and velocities, a motion prediction module predicts future positions of the agents. The corresponding SDF sequence is computed and fed into the hypernetwork and MPC planner. The hypernetwork parametrizes the main network, which approximates the residual component of the terminal constraint.}
    \label{fig:full_diag}
\end{figure}

There are multiple options to select a non-negative activation function at the output of the main network. For example, one can use the ReLU function defined as ${ReLU(x) = \max(0, x)}$. However, its gradient is equal to zero for all negative inputs, which negatively impacts the learning process, causing slow convergence or even model collapse. To overcome this, we use the ELU function defined as
\begin{equation} \label{eq:elu}
ELU(x) =
\begin{cases}
    x, & \text{if } x > 0 \\
    \exp(x) - 1, & \text{if } x \leq 0
\end{cases}
\end{equation}
with a positive unit offset, i.e. we apply $ELU(x) + 1$ at the output of the main network, which guarantees non-negative residual and non-zero gradient for all $x$. In summary, if we denote the main network as $\Phi_{\Theta}(x)$, the value function approximation is defined as
\begin{equation} \label{eq:v_hat}
\begin{aligned} 
    \hat{V}_{k+N}(x) &= F_{k+N}(x) - \hat{R}_{k+N}(x), \\
    \hat{R}_{k+N}(x) &= ELU(\Phi_{\Theta}(x)) + 1.
\end{aligned}    
\end{equation}
Finally, the terminal constraint function is defined as $h_{k+N}(x) = \hat{V}_{k+N}(x)$, and the proposed method is named Residual Neural Terminal Constraint MPC (RNTC-MPC). A simplified diagram is illustrated in \cref{fig:full_diag}.

\textbf{Training procedure} As indicated above, to estimate the value function at the end of the MPC horizon, the model needs to reason about the motion of the locally observed agents after the end of the horizon forward in time. However, from the model's perspective, the task is to estimate the HJ value function based on a sequence of SDFs and the particular point in time when the MPC horizon ends is irrelevant. This means that we can train the model in a simple supervised fashion where the input is an SDF sequence and the target output is the true value function, regardless of when the inference is going to happen.

\begin{figure}[tb]
    \centering
    \includegraphics[width=1\linewidth]{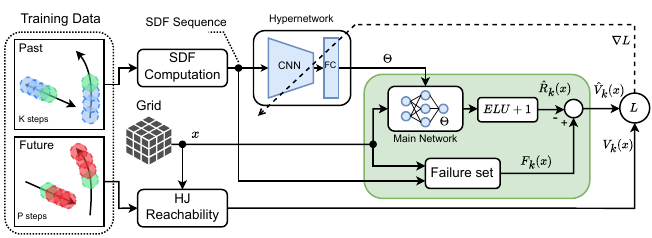}
    \caption{Training diagram of the proposed method. The training data is split into future and past parts of the observed trajectories. The past part is used to compute the SDF sequence, which is fed into the hypernetwork to parametrize the main network. The future part is used to compute the true HJ value function. The predicted value function is obtained as the difference between the numerically computed SDF at the current time and the neural residual estimate.}  
    \label{fig:training_diag}
\end{figure}

To train the model, we start with a dataset of locally observed trajectories of surrounding agents, which can be synthesized in simulations or recorded in the real world. The trajectories can be represented in discrete time (sequence of positions and the corresponding time stamps) or in a continuous domain (continuous function of time). The trajectories are split at the current time $k$ into two components - the future and the past components. The past $K$ time steps are used to compute the SDF sequence that is fed into the hypernetwork, and the future $P$ time steps are used to compute the corresponding true value function $V_k(x)$ using HJ reachability analysis, where $P$ is chosen large enough to ensure that the numerical computation of the value function at time step $k$ converges to $V_k(x)$. The output of the hypernetwork parametrizes the main network at the current time $k$, which renders an approximated residual function $\hat{R}_k(x)$ over the state-space grid. In parallel, the SDF at time $k$ is evaluated over the same grid of states, and the approximated value function $\hat{V}_k(x)$ is computed according to \cref{eq:v_hat}. In the end, the predicted $\hat{V}_k(x)$ is compared with the true $V_k(x)$ using a loss function $L$ and the parameters of the hypernetworks are updated via the stochastic gradient descent (SGD) method based on the loss function gradients $\nabla L$. The training process is visualized in \cref{fig:training_diag}.

For a more robust convergence, we propose to use a combined MSE and exponential (CME) loss, which is defined as:
\begin{equation} \label{eq:cme_loss}
    L = \mathbb{E}_{x\sim Data} \left[ \gamma(V(x) - \hat{V}(x))^2 + (1-\gamma)\exp\left(-V(x)\hat{V}(x)\right) \right],
\end{equation}
where $\gamma \in [0, 1]$ is a hyperparameter that defines the tradeoff between the two components. More detailed analysis of the proposed loss function is presented in \cref{apx:loss_func}.


\section{Experimental Results}
\label{sec:result}

\textbf{Robot Model} In our experiments, we consider the kinematic unicycle model of a mobile robot, which is a common choice in practice. The state vector $x = \left[ x_p, y_p, \theta \right]^\top$  includes position $(x_p, y_p)$ and heading angle $\theta$, while the control inputs are linear velocity $v$ and angular velocity $\omega$. The model is described by the following differential equation:
\begin{equation} \label{eq:kin_unicycle}
    \dot{x} = \left[ v \cdot\cos(\theta), \, v \cdot\sin(\theta), \, \omega \right]^\top,
\end{equation}
with control limits $v \in [-0.5, 0.5]$ m/s and $\omega \in [-0.5, 0.5]$ rad/s.

\textbf{Model Training} To train the model, we first create a dataset of obstacle trajectories in a simulation environment using a constant velocity model. We assume that the perception field of the robot is limited to a square area of 8$\times$8 m centered at the robot's position. Position and velocity vectors of the local obstacles are randomly selected, and the trajectories are simulated forward in time. Next, as explained in \cref{sec:methodology}, we split every trajectory into past and future segments and compute SDF sequences and the true value functions. As a hypernetwork, we use a deep CNN model that takes an SDF sequence as input and outputs the main network parameters. The main network is a simple MLP model with three inputs and a single output with $ELU + 1$ activation function, which estimates residual over the state space grid. More details can be found in \cref{apx:model_train}.

\textbf{Simulation Experiments} We examine the proposed RNTC-MPC framework and our baselines in the Gazebo simulation environment. The task of the robot is to navigate from the initial position to the goal position while avoiding collisions with the moving obstacles based only on the local observations. It is assumed that the accurate estimates of the positions and velocities of the obstacles within the sensing range are accessible. We generate 100 random initializations of the moving obstacles and reuse those initializations for different methods and horizon lengths. Implementation details can be found in \cref{apx:sim_exp}. We compare our method against three main baselines: SDF-MPC, DCBF-MPC \cite{zeng_safety-critical_2021}, and VO-MPC \cite{piccinelli_mpc_2023}. Additionally, as part of an ablation study, we compare it to NTC-MPC \cite{derajic_learning_2025} (with our modification to handle dynamic environments) to show the effect of learning residual instead of the value function directly. All methods use a constant velocity model to predict the motion of the moving obstacles.

\begin{figure}[htb]
  \centering
  \begin{minipage}[b]{0.48\textwidth}
    \centering
    \includegraphics[width=\textwidth]{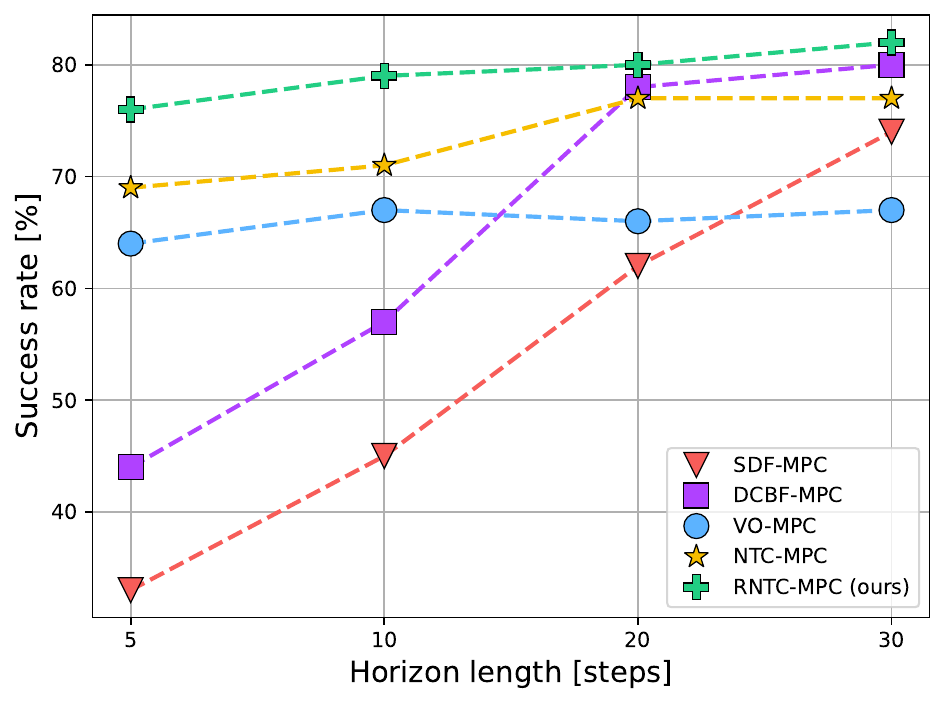}
  \end{minipage}
  \hfill
  \begin{minipage}[b]{0.48\textwidth}
    \centering
    \includegraphics[width=\textwidth]{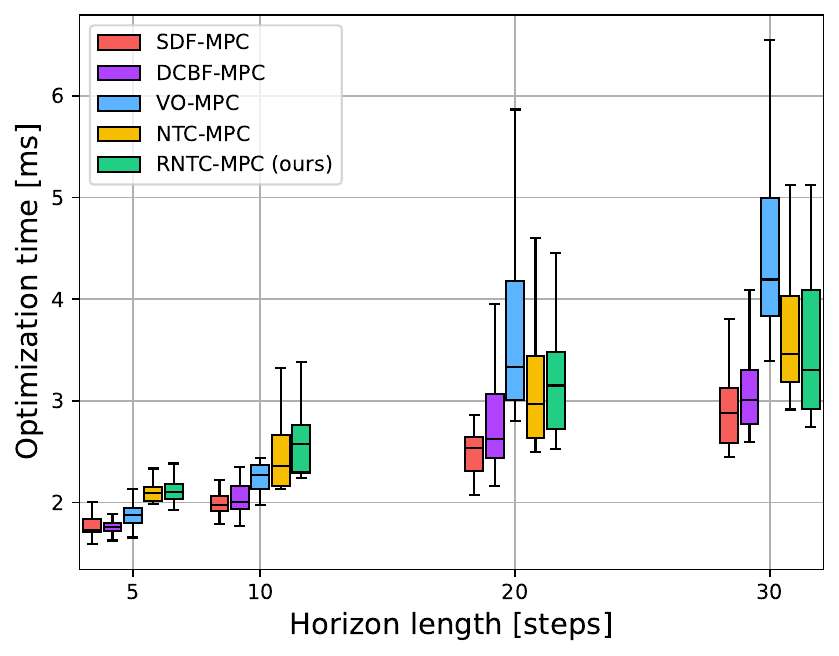}
  \end{minipage}
  \caption{(Left) Success rates of the proposed method and the baselines for different MPC horizon lengths. Our method achieves the highest success rate regardless of the horizon length. (Right) Mean computation time. Our method achieves real-time performance comparable to the baselines. }
  \label{fig:sr_and_opt_time}
\end{figure}

During the experiments, we consider standard metrics for mobile robot navigation. The first metric is success rate, i.e., the percentage of experiments completed without collisions, which is presented in \cref{fig:sr_and_opt_time} (left). The results show a higher success rate of the proposed RNTC-MPC method compared to different baselines, especially for short MPC horizon lengths. \cref{fig:sr_and_opt_time} (right) shows the mean optimization time for the MPC methods, illustrating that the proposed method achieves comparable real-time performance. 

\begin{figure}[htb]
  \centering
  \begin{minipage}[b]{0.48\textwidth}
    \centering
    \includegraphics[width=\textwidth]{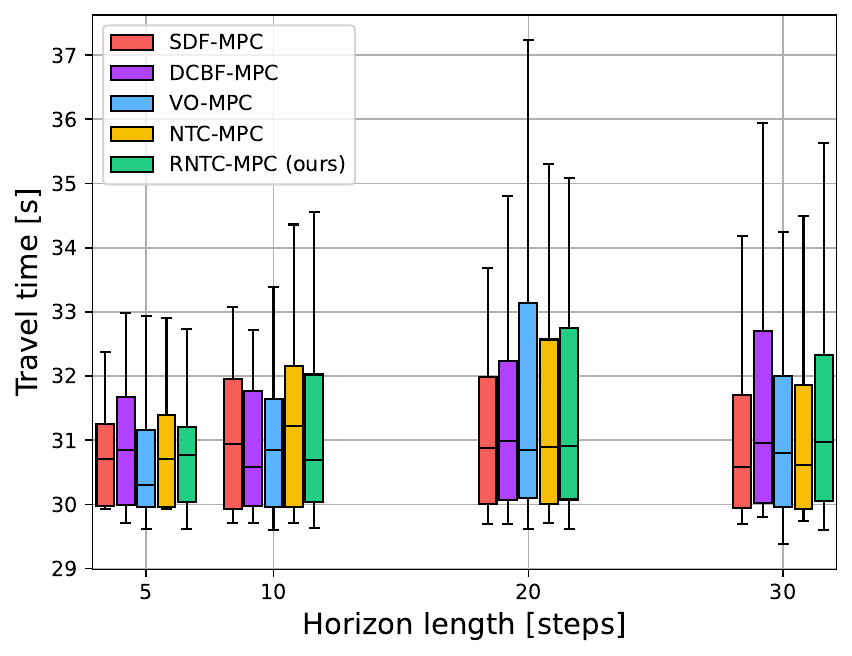}
  \end{minipage}
  \hfill
  \begin{minipage}[b]{0.48\textwidth}
    \centering
    \includegraphics[width=\textwidth]{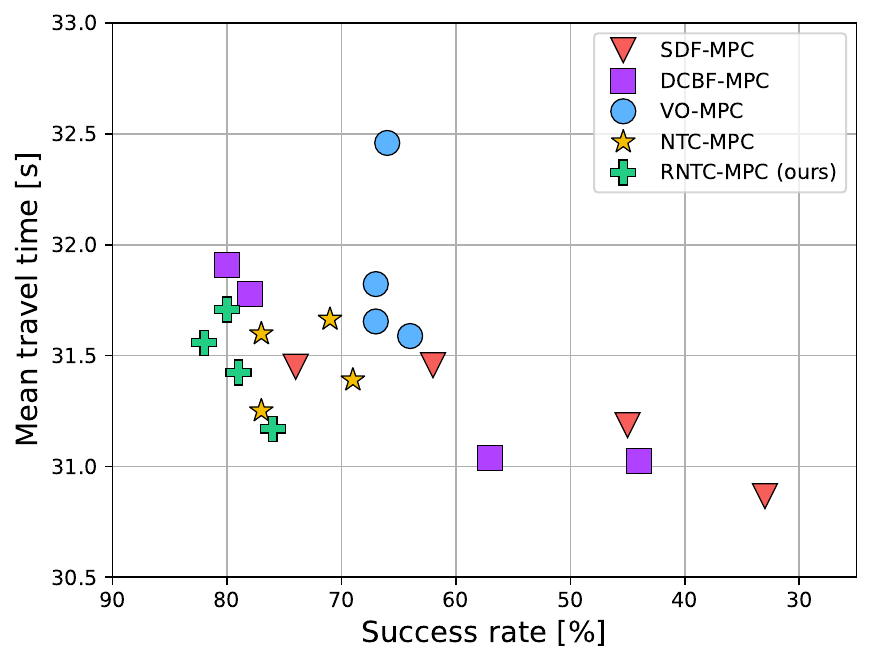}
  \end{minipage}
  \caption{(Left) Mean travel times during navigation between initial and goal positions. (Right) Trade-off between success rate and mean travel time. Our method achieves the best compromise between the two objectives.}
  \label{fig:travel_time_and_pareto_front}
\end{figure}

\cref{fig:travel_time_and_pareto_front} (left) shows the travel time needed for the robot to move from the initial to the goal position, suggesting similar results for all methods. However, \cref{fig:travel_time_and_pareto_front} (right) illustrates the trade-off between the two objectives - success rate and mean travel time, which clearly indicates that our method achieves the best compromise between the two objectives.  

\begin{figure}[htb]
  \centering
  \begin{minipage}[b]{0.48\textwidth}
    \centering
    \includegraphics[width=\textwidth]{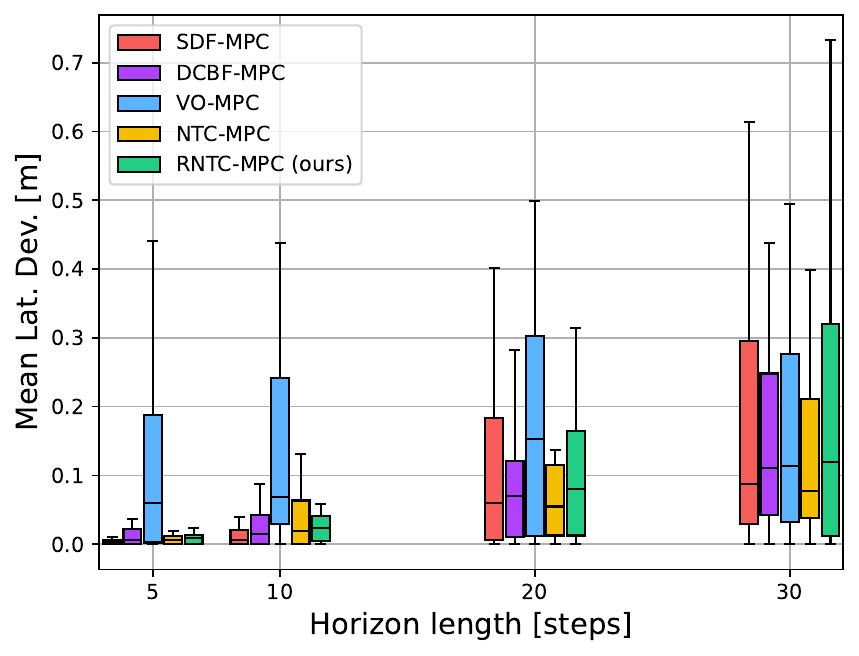}
  \end{minipage}
  \hfill
  \begin{minipage}[b]{0.48\textwidth}
    \centering
    \includegraphics[width=\textwidth]{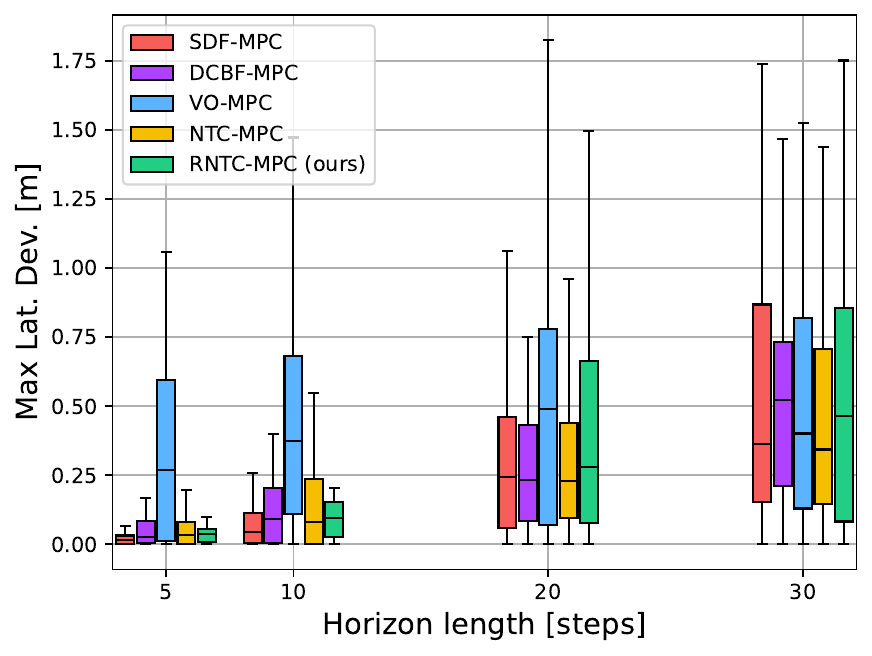}
  \end{minipage}
  \caption{(Left) Mean lateral deviation. (Right) Maximal lateral deviation.}
  \label{fig:lat_dev}
\end{figure}

Additionally, we examine lateral deviation $d$ from the reference path, which is defined as the closest distance from the robot position to the reference path. This metric is particularly important in applications where the path following is crucial for successful operation (e.g. cleaning robot). The mean lateral deviation $d_{mean}$ across experiments is shown in \cref{fig:lat_dev} (left) and the maximal lateral deviation $d_{max}$ is shown in \cref{fig:lat_dev} (right). The results show that the VO-MPC method has the lowest performance overall, while our method is comparable to other baselines.

\textbf{Hardware Experiments} In addition to simulations, we assess the proposed method on a real robot and compare it to the main baselines: SDF-MPC, DCBF-MPC, and VO-MPC. We use a mobile robot equipped with 3D LiDAR and stereo cameras for our experiments. All computation and data processing is performed onboard. Similar to the simulation experiments, the task is to move from the initial position to the goal position while avoiding moving obstacles. We perform 10 trials for each method in scenarios where pedestrians intentionally intersect robot motion. The results are presented in \cref{tab:hardware_experiment}, illustrating 30\% higher success rate compared to the baselines, while achieving the lowest travel time. Additional details and qualitative results for hardware experiments are presented in \cref{apx:hardw_exp} and the supplemental video\footnote{\url{https://www.youtube.com/watch?v=eyaAYIzgBpI}}. The discrepancy in results between simulations and hardware experiments can be explained by the lower number of trials in the hardware experiments.

\begin{table}[tbh]
\centering
\small
\caption{Results from the hardware experiments for MPC prediction horizon of 10 steps (1 s).}
\label{tab:hardware_experiment}
\begin{tabular}{lccccc}
\hline
Method & Success rate [\%] & $d_{mean}$ [m] & $d_{max}$ [m] & Opt. time [ms] & Travel time [s] \\
\hline
SDF-MPC         & 40.0 & 0.03 {\scriptsize$\pm$ 0.01} & 0.08 {\scriptsize$\pm$ 0.01} & 2.93 {\scriptsize$\pm$ 0.25} & 13.03 {\scriptsize$\pm$ 1.73} \\
DCBF-MPC        & 70.0 & 0.11 {\scriptsize$\pm$ 0.06} & 0.36 {\scriptsize$\pm$ 0.22} & 3.09 {\scriptsize$\pm$ 0.38} & 13.98 {\scriptsize$\pm$ 0.93} \\
VO-MPC          & 70.0 & 0.19 {\scriptsize$\pm$ 0.13} & 0.53 {\scriptsize$\pm$ 0.33} & 6.94 {\scriptsize$\pm$ 2.61} & 14.67 {\scriptsize$\pm$ 1.55} \\
RNTC-MPC (ours) & 100.0 & 0.04 {\scriptsize$\pm$ 0.02} & 0.10 {\scriptsize$\pm$ 0.08} & 4.20 {\scriptsize$\pm$ 0.56} & 12.76 {\scriptsize$\pm$ 0.89} \\
\hline
\end{tabular}
\end{table}


\section{Conclusion}
\label{sec:conclusion}

This paper proposes a learning-based method for maximal safe set estimation in dynamic environments. The set is estimated based on the predicted motion of the moving obstacles and used as the terminal constraint for the MPC local planner. We exploit the property that the HJ value function, which implicitly characterizes the safe set, is less than or equal to the corresponding SDF. Our method approximates only the residual part of the value function, which is subtracted from the numerically computed SDF. By using a neural model with non-negative output, it is guaranteed that the resulting value function approximation is at least as safe as the SDF for all states.

The results, both in simulation and hardware experiments, illustrate the advantage of our approach compared to all baselines. The proposed RNTC-MPC method achieves higher success rates,  especially for short MPC horizon lengths, while achieving comparable real-time performance and motion quality. In future work, we will mainly focus on improving the scalability to higher-dimensional systems.

\clearpage

\section{Limitations}
\label{sec:limitations}

This section outlines several areas where the RNTC-MPC method could be further developed or improved. One technical challenge arises during the training phase, where we rely on numerical HJ reachability tools to generate ground truth data. This approach is known to be computationally intensive, particularly for systems with higher-dimensional state spaces (typically beyond six dimensions), due to the curse of dimensionality. To address this, future work might explore self-supervised training strategies, such as those proposed in \cite{bansal_deepreach_2021}.

Another consideration involves the strength of formal guarantees on safe set approximation error. Even though the proposed method makes a significant step in this direction, more advantageous safety guarantees are desirable. In that regard, one could try to exploit the mathematical properties of the HJ value function further, or to employ regularized models for function approximation with statistical guarantees presented in \cite{taheri_statistical_2021}.

Finally, in this work, we assume constant velocity for obstacles, which is a reasonable approximation in some settings, especially when using short prediction horizons, but may not capture the complexity of more dynamic environments. Even though this assumption is not an explicit limitation of the proposed method, a more general approach would be to consider a distribution of possible future trajectories and compute a stochastic or robust terminal set. In that case, the constraint could be reformulated in a form similar to \cite{zhang_trajectory_2021}. Also, other rule-based or learning-based motion models (see \cite{korbmacher_review_2022} for an overview) could be used.


\acknowledgments{This work was partially funded by the German Federal Ministry for Economic Affairs and Climate Action within the project \textit{nxtAIM} and Continental Automotive GmbH.}


\bibliography{references}

\newpage

\appendix

\section{Ablation Study on Model Architecture}
\label{apx:ablation_study}

In this section, we examine the advantage of the proposed method through an ablation study. We compare the proposed RNTC-MPC method to the NTC-MPC method \cite{derajic_learning_2025}, which uses the main network to approximate the HJ value function directly, but with modifications to handle dynamic environments. Therefore, here we illustrate the benefits of approximating only the non-negative residual, which is then subtracted from the numerically computed SDF in order to obtain HJ value function approximation.

\begin{figure}[tbh]
  \centering
  \begin{minipage}[b]{0.48\textwidth}
    \centering
    \includegraphics[width=\textwidth]{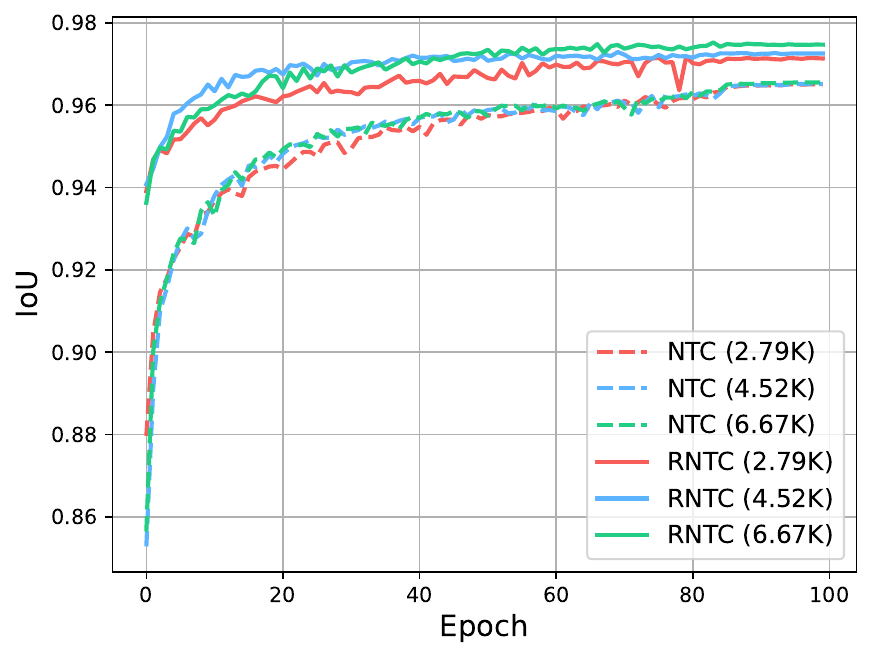}
  \end{minipage}
  \hfill
  \begin{minipage}[b]{0.48\textwidth}
    \centering
    \includegraphics[width=\textwidth]{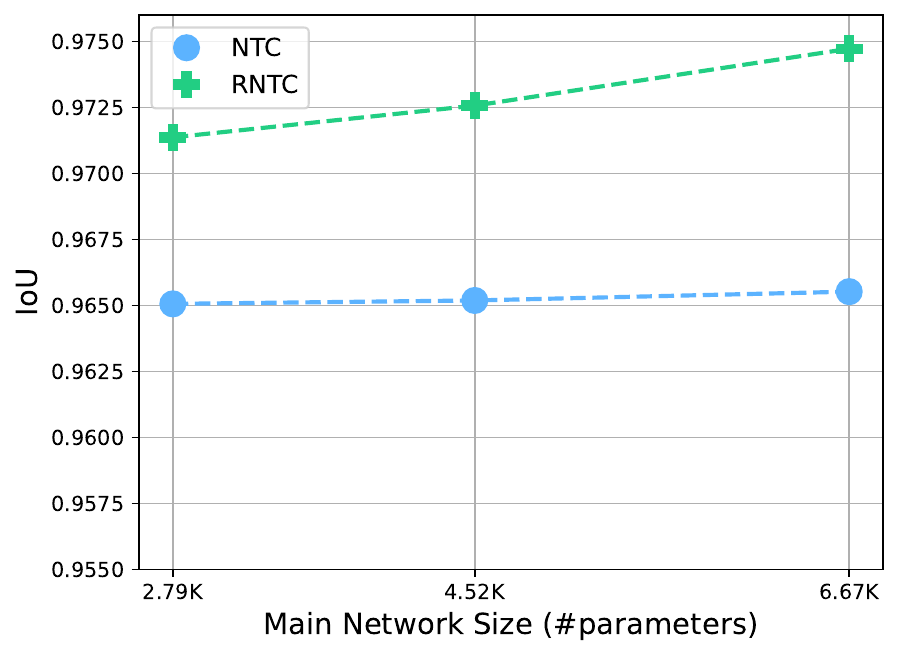}
  \end{minipage}
  \caption{(Left) IoU metric on the validation dataset during training for three different sizes of the main network. (Right) The final IoU values after 100 epochs of training for three different sizes of the main network.}
  \label{fig:iou_ntc_rntc}
\end{figure}

The first benefit can be observed during the training phase - the model trained in the RNTC method converges faster and achieves higher approximation accuracy for the same number of parameters. In \cref{fig:iou_ntc_rntc} (left), we show the Intersection over Union (IoU) metric for the recovered safe set on the validation dataset during training for three different sizes of the main network. The IoU metric, in addition to the loss function itself, gives us information about the result quality, i.e., how well the model separates safe from unsafe regions. The achieved IoU metric for the RNTC method after 10 epochs is approximately equal to the IoU achieved by the NTC method after 50 epochs, which suggests five times faster convergence. Also, in \cref{fig:iou_ntc_rntc} (right), we show the final IoU value after 100 epochs for three different sizes of the main network, and the RNTC approach achieves higher values in all cases. Those observations can be explained by significant inductive bias in the RNTC method, i.e. the model needs to learn the residual part of the HJ value function, and the model is additionally constrained to output only non-negative values. Conversely, the NTC method learns the value function from scratch. 

\begin{figure}[tbh]
    \centering
    \includegraphics[width=\linewidth]{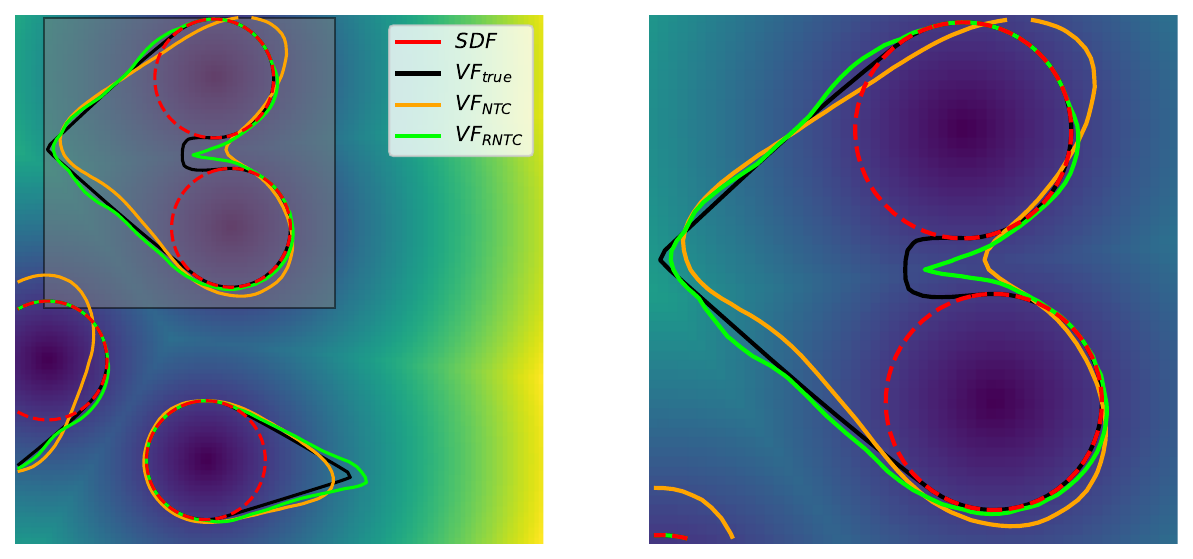}
    \includegraphics[width=\linewidth]{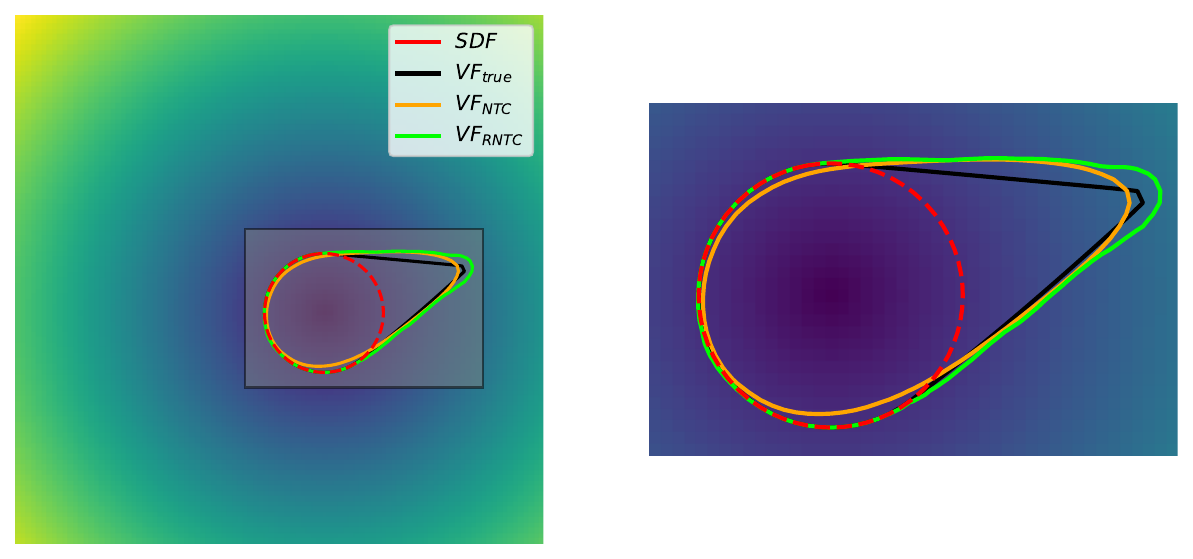}
    \caption{Two samples of HJ value function slices for $\theta=0$ rad (right - full slice; left - zoomed slice). The results illustrate that the value function approximation in the RNTC method classifies states as unsafe if they are unsafe according to the SDF, which is not true for the NTC method.}  
    \label{fig:ntc_vs_rntc}
\end{figure}

In \cref{fig:ntc_vs_rntc} we qualitatively illustrate the advantage of the RNTC compared to the NTC approach. The two HJ value function slices visualize the unsafe regions predicted by RNTC and NTC methods, the true unsafe region, and the unsafe region predicted by the corresponding SDF. It can be seen that the RNTC method classifies all states as unsafe that are unsafe according to the SDF. On the other side, the NTC method assigns positive values to some states that are unsafe according to SDF (occupied space), which eventually can lead to unsafe actions.

\section{CME Loss Function Analysis}
\label{apx:loss_func}

\begin{figure}[tbh]
  \centering
  \begin{minipage}[b]{0.48\textwidth}
    \centering
    \includegraphics[width=\textwidth]{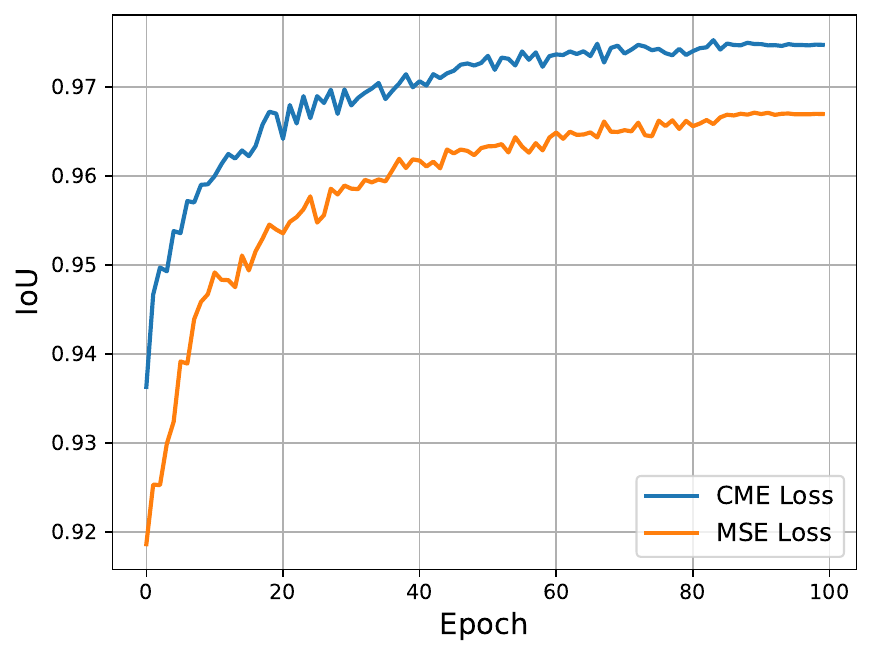}
  \end{minipage}
  \hfill
  \begin{minipage}[b]{0.48\textwidth}
    \centering
    \includegraphics[width=\textwidth]{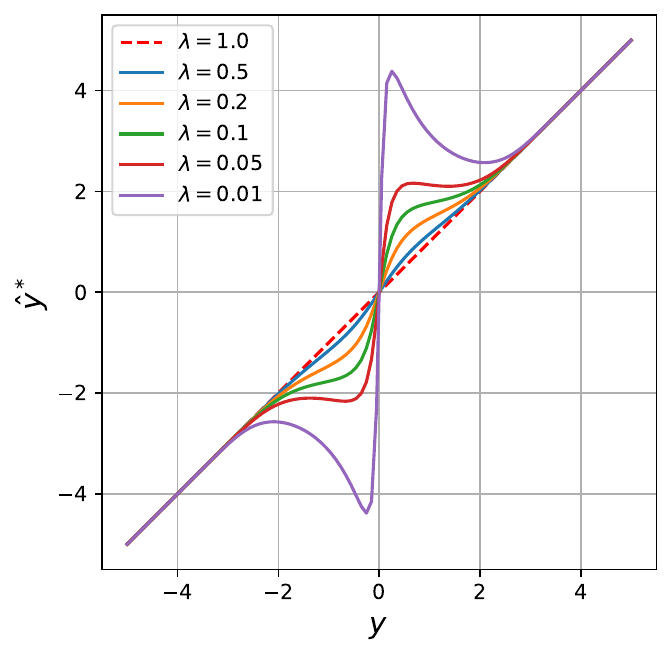}
  \end{minipage}
  \caption{(Left) IoU metric evaluated on the validation dataset during training for the standard MSE loss and the proposed CME loss. (Right) Relation between optimal prediction $\hat{y}^*$ and target output $y$ for different values of hyperparameter $\gamma$.}  
    \label{fig:mse_vs_cme_loss _and_yhat_vs_y}
\end{figure}

The main motivation for the proposed hybrid CME loss function is to achieve faster convergence during the training process and better final performance. We compare the proposed loss defined in \cref{eq:cme_loss} with the standard MSE loss function used in regression tasks. We train the same model under the same settings and on the same dataset. \cref{fig:mse_vs_cme_loss _and_yhat_vs_y} (left) shows IoU metric on the validation dataset during training. The proposed CME loss converges significantly faster, and the final IoU value at the end of training is higher\footnote{To prevent numerical issues in the first epoch caused by the exponential term in the CME loss, one can pretrain the model using the MSE loss for a single epoch.}

To analyze the hybrid CME loss function defined by \cref{eq:cme_loss} in more detail, we consider a general regression problem with target outputs $y$ and predicted outputs $\hat{y}$. We assume that the target outputs $y$ represent samples of some value function whose zero-level set separates safe and unsafe states in the system state space. Therefore, the optimal predictions $\hat{y}^*$ must have the same zero-level set as the target outputs $y$. To prove this for the proposed CME loss function, we need to find a relation between the loss function optimum and target values. If we consider a dataset with a single training pair, the loss function is
\begin{equation}
    L = \gamma (y - \hat{y})^2 + (1 - \gamma) \exp \left( -y \hat{y} \right).
\end{equation}
Since $L$ is a linear combination of two convex functions with non-negative coefficients, it is also a convex function and we can find the global optimum by simply finding a stationary point.  The derivative of $L$ w.r.t. $\hat{y}$ is
\begin{equation}
    \frac{\partial L}{\partial \hat{y}} = -2 \gamma (y - \hat{y}) - y(1 - \gamma) \exp \left( -y \hat{y} \right),
\end{equation}
and optimal prediction $\hat{y}^*$ that minimizes loss is a root of the following transcendental equation
\begin{equation}
    -2 \gamma (y - \hat{y}) - y(1 - \gamma) \exp \left( -y \hat{y} \right) = 0.
\end{equation}
The optimal solution can be written as
\begin{equation}
    \hat{y}^* = y + \frac{1}{y}W \left( \frac{(1 - \gamma)y^2}{2\gamma \exp(y^2)} \right),
\end{equation}
where $W(\cdot)$ is so-called Lambert W function \cite{corless_lambertw_1996}. It is clear that the optimal model output $\hat{y}^*$ is different from the corresponding target value $y$. However, based on the W function properties \cite{corless_lambertw_1996} and using L'Hôpital's rule as $y \to 0$, one can show that the $\hat{y}^*$ and $y$ have the same sign and zero-level set.

In \cref{fig:mse_vs_cme_loss _and_yhat_vs_y} (left), we show the relation between optimal prediction $\hat{y}^*$ and target output $y$ for different $\gamma$. For $\gamma = 1$, the loss function is equal to MSE and the optimal prediction is equal to the target value. As we decrease $\gamma$, the relation becomes more nonlinear near the zero target value. Nonetheless, the zero value is mapped to zero, and the predicted sign is the same as for the target value, which confirms the theoretical discussion above. However, for relatively small values of $\gamma$, the exponential component of the CME loss might cause numerical issues due to random initialization of model weights. A simple solution is to use the MSE loss in the first epoch of the training, and then continue with the CME loss. 

\section{Model Training}
\label{apx:model_train}

To train the model, we generate a dataset of obstacle trajectories using the constant velocity model for obstacle motion. It is important to mention that the dataset of trajectories can be generated using any other motion model or by collecting real-world data. The trajectories are split at some point in time into past and future segments, which are then used to calculate corresponding SDF sequences and true HJ value functions. In this paper, we assume that the robot can sense moving obstacles within a square region of size 8$\times$8 meters centered at the robot's location. The positions and velocity vectors of the local obstacles are randomly sampled within this region, and the trajectories are then simulated forward in time. We assume that the maximal number of obstacles is 4 and that they can move in any direction with a maximal speed of up to 1 m/s. In total, we generate 40.000 training pairs, i.e., pairs of SDF sequences and the corresponding HJ value function.

For HJ value function computation, we use the \textit{hj\_reachability}\footnote{\url{https://github.com/StanfordASL/hj_reachability}} Python library with modifications to handle time-varying failure sets. The value function is computed over a discrete state space grid of size 100$\times$100$\times$30. The computation is performed on a workstation with two NVIDIA RTX3090 GPUs and the process took approximately 20h to complete.

The hypernetwork, the main network, and the training process are implemented using the PyTorch framework. The hypernetwork is implemented as a standard CNN model that takes as input an SDF sequence and outputs the main network parameters $\Theta$. In our concrete case, the SDF sequence contains only 2 SDFs (at the current time and 0.4 s into the past) because we assumed the constant velocity model, and two SDFs contain enough information to forecast the future motion. The complete hypernetwork architecture is provided in \cref{lst:hypernet_summary}, resulting in $\sim9.37 \cdot 10^6$ trainable parameters. The main network is implemented as an MLP model with three input neurons (size of the state vector) and a single output with $ELU + 1$ activation function to guarantee a non-negative estimated residual. The details are presented in \cref{lst:main_net_summary}, and the total number of parameters is 4519. The residual is subtracted for the SDF at the current time to estimate the HJ value function that corresponds to the input SDF sequence. Then, the predicted value function is compared to the true value function via the loss function \cref{eq:cme_loss} (for $\gamma = 0.1$, selected via grid search) and the weights of the hypernetwork are updated using mini-batch gradient descent. We train the model for 100 epochs using Adam optimizer with an initial learning rate of 0.0001 and a batch size of 40. The learning rate is reduced by a factor of 10 at epochs 85 and 95. The training process takes approximately 15h on an NVIDIA RTX3090 GPU.

\begin{lstlisting}[language=Python, caption=Hypernetwork model summary., label={lst:hypernet_summary}]
Hypernetwork(
  (backbone): Sequential(
    (0): Conv2d(2, 16, kernel_size=(5, 5), stride=(1, 1), padding=valid)
    (1): ReLU()
    (2): MaxPool2d(kernel_size=(2, 2), stride=(2, 2), padding=0, dilation=1, ceil_mode=False)
    (3): Conv2d(16, 32, kernel_size=(5, 5), stride=(1, 1), padding=valid)
    (4): ReLU()
    (5): MaxPool2d(kernel_size=(2, 2), stride=(2, 2), padding=0, dilation=1, ceil_mode=False)
    (6): Conv2d(32, 64, kernel_size=(3, 3), stride=(1, 1), padding=valid)
    (7): ReLU()
    (8): MaxPool2d(kernel_size=(2, 2), stride=(2, 2), padding=0, dilation=1, ceil_mode=False)
    (9): Conv2d(64, 128, kernel_size=(3, 3), stride=(1, 1), padding=valid)
    (10): ReLU()
    (11): MaxPool2d(kernel_size=(2, 2), stride=(2, 2), padding=0, dilation=1, ceil_mode=False)
    (12): Flatten(start_dim=1, end_dim=-1)
  )
  (head): Sequential(
    (0): Linear(in_features=2048, out_features=4519, bias=True)
  )
)
\end{lstlisting}

\begin{lstlisting}[language=Python, caption=Main network model summary., label={lst:main_net_summary}]
MainNetwork(
  (model): Sequential(
    (layer_0): Linear(in_features=3, out_features=36, bias=True)
    (activation_0): Sin()
    (layer_1): Linear(in_features=36, out_features=36, bias=True)
    (activation_1): Sin()
    (layer_2): Linear(in_features=36, out_features=36, bias=True)
    (activation_2): Sin()
    (layer_3): Linear(in_features=36, out_features=18, bias=True)
    (activation_3): SELU()
    (layer_4): Linear(in_features=18, out_features=18, bias=True)
    (activation_4): SELU()
    (layer_5): Linear(in_features=18, out_features=18, bias=True)
    (activation_5): SELU()
    (layer_6): Linear(in_features=18, out_features=9, bias=True)
    (activation_6): SELU()
    (layer_7): Linear(in_features=9, out_features=9, bias=True)
    (activation_7): SELU()
    (layer_8): Linear(in_features=9, out_features=9, bias=True)
    (activation_8): SELU()
    (output_layer): Linear(in_features=9, out_features=1, bias=True)
  )
)
\end{lstlisting}

\section{Simulation Experiments}
\label{apx:sim_exp}

\begin{figure}[tbh]
    \centering
    \includegraphics[width=0.8\linewidth]{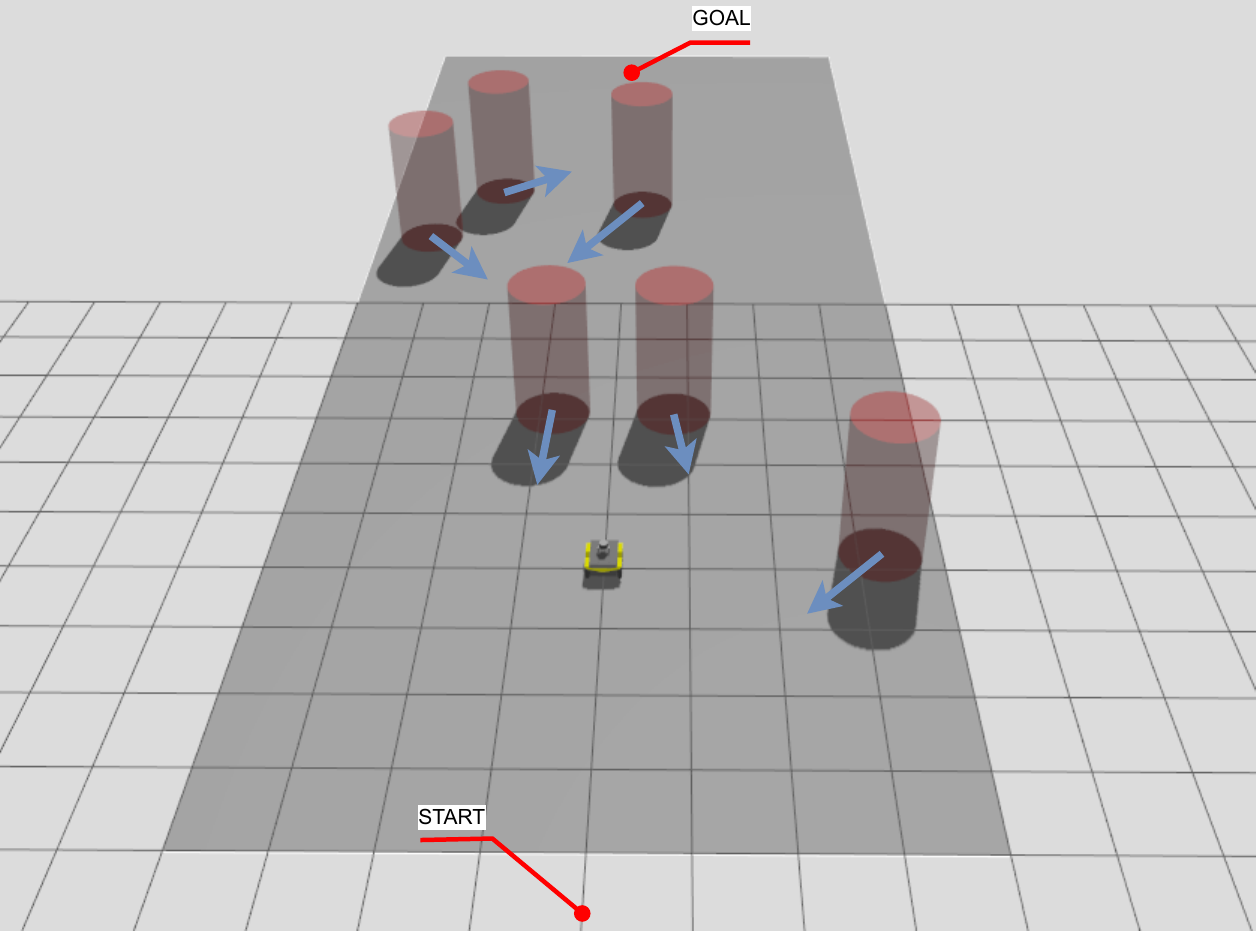}
    \caption{Visualization of the Gazebo environment used in simulation experiments. The environment includes six obstacles with randomly initialized positions and velocity vectors. The task is to navigate from that start position to the goal position while avoiding collisions.}  
    \label{fig:gazebo_env}
\end{figure}

\begin{figure}[tbh]
    \centering
    \includegraphics[width=1.0\linewidth]{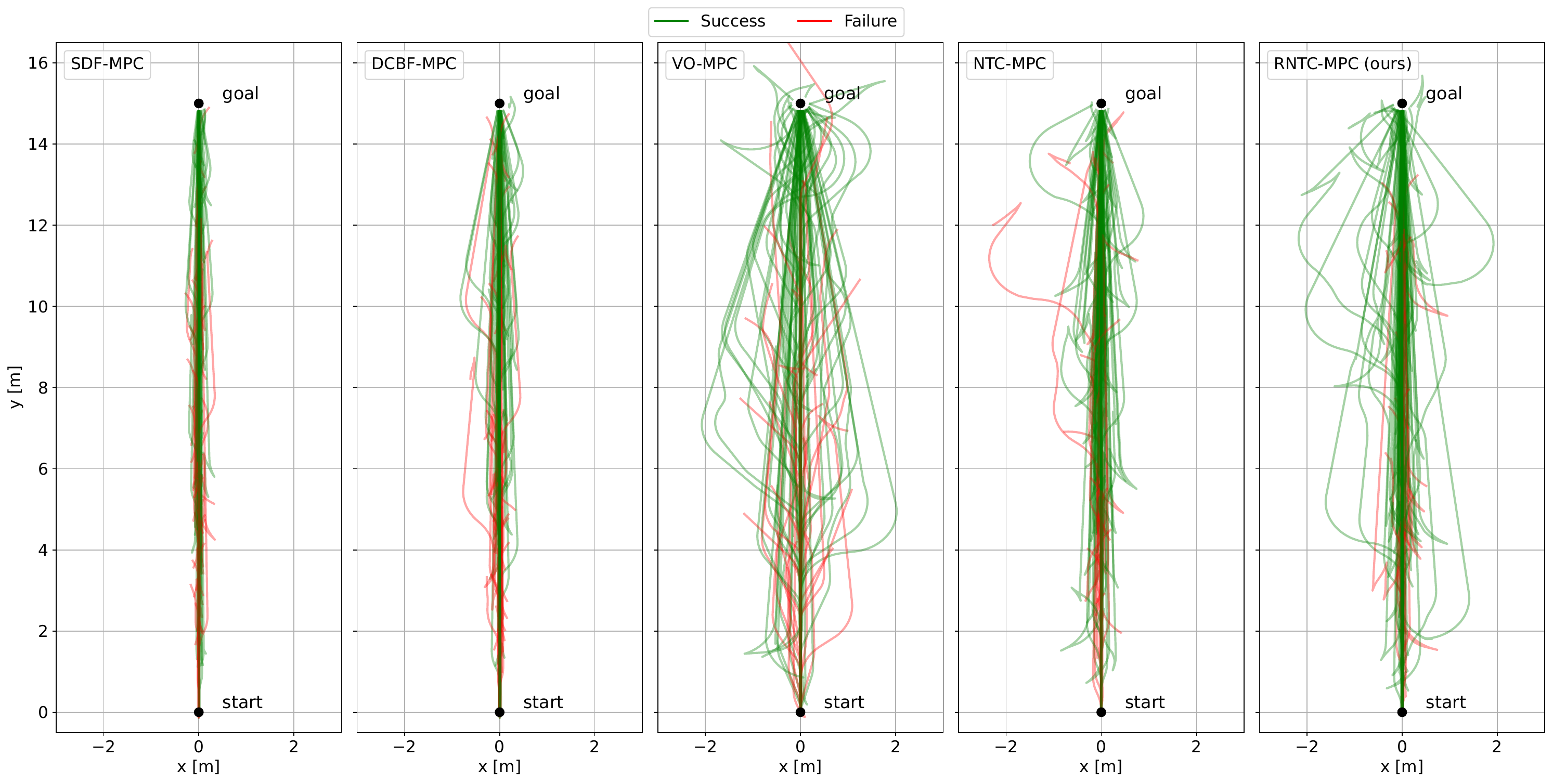}
    \caption{Visualization of the motion executed by the robot during simulation experiments in all trials for different MPC types with the prediction horizon of 10 steps (1 s).}  
    \label{fig:motion_sim_exp}
\end{figure}

The simulation experiments are performed using the Gazebo 3D simulator. The task for the robot is to move from the initial state to the goal state while avoiding collisions with moving obstacles. We generate 100 different scenarios with random initialization of positions and velocities for six moving obstacles within the corridor that the robot needs to pass through. The obstacles move with constant speed and bounce back when they reach the edge of the corridor, so that all obstacles stay within the corridor all the time. The testing environment is shown in \cref{fig:gazebo_env} and the motion executed by the robot for all MPC types with the prediction horizon of 10 steps (1 s) is visualized in \cref{fig:motion_sim_exp}. 

The MPC local planner is implemented using the CasADi framework \cite{andersson_casadi_2019} as a ROS 2 package with the control frequency of 20 Hz and sampling time $\delta t=0.1$ s. To improve the real-time performance of the planner, we utilize CasADi's feature to autogenerate C code and compile it before deployment, and additionally, we use the high-performance linear solver MA57 \cite{duff_ma57---code_2004}, which significantly improves computational time. The hypernetwork is implemented as a PyTorch model, the same as during the training phase, and achieves an inference speed of $\sim$2 ms on an RTX3090 GPU. However, integrating an MLP model with dynamically updated parameters as the MPC terminal constraint function is not trivial. Our solution is to implement the main network as a custom Python class using the operations provided by CasADi. In this way, the main network is integrated into the computational graph of the optimization problem, just like other functions. Also, the parameters of the main network are declared as parameters of the NLP solver, so they can be changed at every time step (same as the estimated and reference states). 

\section{Hardware Experiments}
\label{apx:hardw_exp}

The hardware experiments are performed using a Continental mobile delivery robot equipped with a 3D LiDAR and stereo cameras. Complete computation and data processing, such as object recognition and tracking, robot localization, MPC optimization, etc., are performed using the onboard computer\footnote{Intel NUC11PHi7 (CPU: 11th Gen Intel® Core™ i7-1165G7 @ 2.80GHz × 8; RAM: 16GB; GPU: NVIDIA GeForce RTX 2060).}. As in the simulation experiments, the MPC planner is implemented using CasADi and deployed as a ROS 2 package. Also, the hypernetwork is deployed as a PyTorch model (with inference speed of $\sim$4.5 ms on an RTX2060 GPU), while the main network is implemented using CasADi functions. The discretization time of the MPC is $\delta t = 0.1$ s and the control frequency is 20~Hz. Motion prediction of moving agents is performed using the constant velocity model.

\begin{figure}[tbh]
    \centering
    \includegraphics[width=1.0\linewidth]{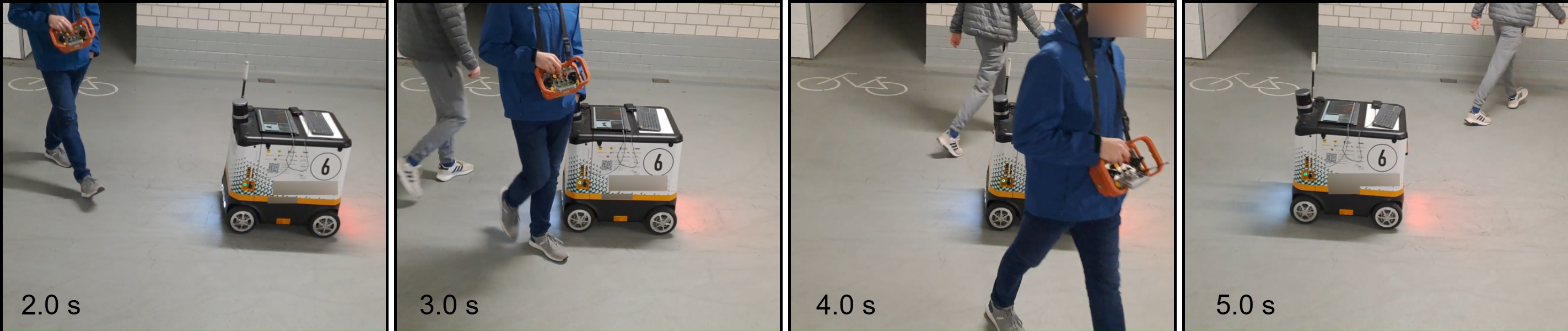}
    \caption{Illustration of a successful collision avoidance of the proposed RNTC-MPC method with the prediction horizon of 10 steps (1 s) during the hardware experiments.}  
    \label{fig:hardware_exp}
\end{figure}

\begin{figure}[tbh]
    \centering
    \includegraphics[width=0.8\linewidth]{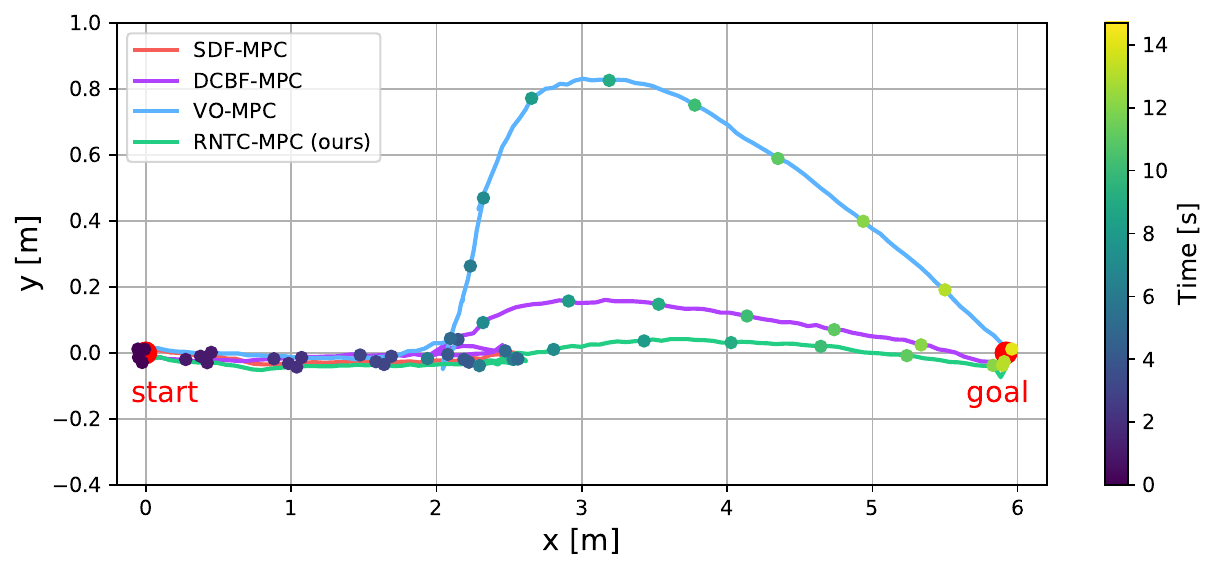}
    \caption{Example motion executed by the robot during hardware experiments. The task is to move from the start position to the goal position, which represents two successive waypoints. The results indicate lower lateral deviation of the proposed RNTC-MPC method compared to the baselines.}  
    \label{fig:sample_traj}
\end{figure}

\begin{figure}[tbh]
    \centering
    \includegraphics[width=1.0\linewidth]{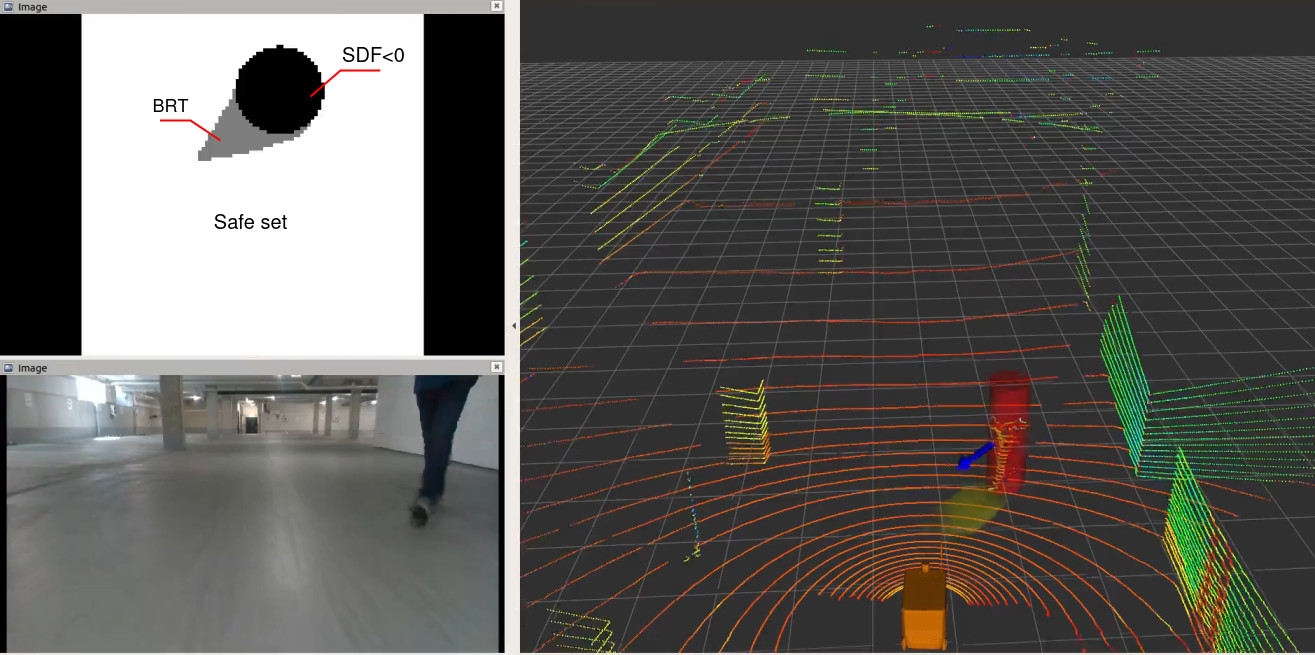}
    \caption{(Left up) Predicted terminal set. (Left down) The front camera view. (Right) Visualization of the detected human and predicted motion.}  
    \label{fig:hardware_exp_screen}
\end{figure}

During the hardware experiments, we conduct 10 trials for each method --- 5 experiments with one pedestrian intersecting robot motion and 5 experiments with two pedestrians. The initial positions and velocities of pedestrians are random. The main difficulty of collision avoidance in those experiments arises from the limited actuation of the robot (relatively low maximal speed compared to moving obstacles), short prediction horizon, and complete onboard processing. In \cref{fig:hardware_exp} we illustrate one example of a successfully avoided collision with a human, even though the MPC operates with a relatively short prediction horizon of 10 steps (1 s). To assess the quality of the executed trajectory, we visualize the motion of the robot for different baselines in \cref{fig:sample_traj}. The results demonstrate significantly lower lateral deviation in the case of the proposed RNTC-MPC method compared to DCBF-MPC and VO-MPC, while SDF-MPC fails to avoid collision. An additional visualization of the robot's perception and estimated terminal set is presented in \cref{fig:hardware_exp_screen}.

\end{document}